\documentclass[twoside]{article}
\usepackage{etoolbox}
\newtoggle{LONG_VERSION}
\toggletrue{LONG_VERSION}
\usepackage{arxiv}

\usepackage[round]{natbib}

\usepackage[utf8]{inputenc} 
\usepackage[T1]{fontenc}    
\usepackage{hyperref}       
\usepackage{url}            
\usepackage{booktabs}       
\usepackage{amsfonts}       
\usepackage{nicefrac}       
\usepackage{microtype}      
\usepackage{graphicx}
\usepackage{booktabs} 
\usepackage{comment}
\usepackage{amsfonts}
\usepackage{amsmath}
\usepackage{amssymb}
\usepackage{amsthm}
\usepackage{mathtools}
\usepackage{bm}
\usepackage{paralist}
\usepackage{float}
\usepackage{xcolor}
\usepackage{adjustbox}
\usepackage{wrapfig}
\usepackage{tabularx}
\usepackage{multirow}
\usepackage{subcaption}

\usepackage{hyperref}

\usepackage{algorithm}
\usepackage[noend]{algorithmic}

\newtheorem{theorem}{Theorem}[section]

\newtheorem{proposition}[theorem]{Proposition}

\newtheorem{definition}[theorem]{Definition}

\DeclareMathOperator*{\argmin}{arg\,min} 

\begin{document}

\twocolumn[

\arxivtitle{Enhancing Certifiable Robustness via a Deep Model Ensemble}

\arxivauthor{ Huan Zhang \And Minhao Cheng \And Cho-Jui Hsieh }

\arxivaddress{ UCLA \\ \texttt{huan@huan-zhang.com} \And  UCLA \\ \texttt{mhcheng@ucla.edu} \And UCLA \\ \texttt{chohsieh@cs.ucla.edu}}
]

\begin{abstract}
We propose an algorithm to enhance certified robustness of a deep model ensemble by optimally weighting each base model. Unlike previous works on using ensembles to empirically improve robustness, our algorithm is based on optimizing a guaranteed robustness certificate of neural networks. Our proposed ensemble framework with certified robustness, \textbf{RobBoost}, formulates the optimal model selection and weighting task as an optimization problem on a lower bound of classification margin, which can be efficiently solved using coordinate descent. Experiments show that our algorithm can form a more robust ensemble than naively averaging all available models using robustly trained MNIST or CIFAR base models.
Additionally, our ensemble typically has better accuracy on clean (unperturbed) data.
RobBoost allows us to further improve certified robustness and clean accuracy by creating an ensemble of already certified models.
\iftoggle{LONG_VERSION}{}{
\vspace{0.5em}\\
\textbf{Contributions.} Huan Zhang proposed the idea of robust model ensemble, formulated the problem as an optimization problem and derived the algorithm on this paper. He also conducted experiments and wrote this paper entirely. Minhao Cheng helped on some of the experiments, collected and visualized results. Cho-Jui Hsieh (advisor) gave advice on improving the proposed algorithm and checked the correctness of the algorithm, and also suggested edits on this paper.}
\end{abstract}

\section{Introduction}

The lack of robustness in deep neural networks (DNNs) has
motivated recent research on verifying and improving the robustness of DNN models~\citep{katz2017reluplex,dvijotham2018dual,gehr2018ai,singh2018fast,madry2018towards,raghunathan2018semidefinite,wong2018provable,zhang2018crown}.
Improving the robustness of neural networks, or designing the ``defense'' to adversarial examples, is a challenging problem.
\citet{athalye2018obfuscated,uesato2018adversarial} showed that many proposed defenses are broken and do not significantly increase robustness under adaptive attacks.
So far,  state-of-the-art defense methods include adversarial training~\citep{madry2018towards,sinha2018certifying} and optimizing a certified bound on robustness~\citep{wong2018provable,raghunathan2018certified,wang2018mixtrain,mirman2018differentiable}, but there is still a long way to go to conquer the adversarial example problem. On MNIST at perturbation $\epsilon=0.3$, adversarially trained models~\citep{madry2018towards} are resistant to strong adversarial attacks but cannot be efficiently certified using existing neural network verification techniques. On the other hand, certifiable training methods usually suffer from high clean and verified error; in \citep{wong2018scaling}, the best model achieves 43.1\% verified error and 14.9\% clean error, which is much higher than ordinary MNIST models.



Most of these existing defense methods only focus on improving the robustness of a single model. Traditionally, a model ensemble has been used to improve prediction accuracy of weak models. 
For example, voting or bootstrap aggregating (bagging) can be used to improve prediction accuracy in many occasions.
Furthermore, boosting based algorithms, including AdaBoost~\citep{freund1997decision}, LogitBoost~\citep{friedman2000additive}, gradient boosting~\citep{friedman2001greedy,friedman2002stochastic} and many other variants, are designed to minimize an upper bound (surrogate loss) of classification error, which provably increases the model's accuracy despite the fact that each base model can be very weak. Inspired by the successful story of model ensembles, our question is that if a similar technique can be used to build a robust model with better \emph{provable} robustness via an ensemble of certifiable base models?

Intuitively, attacking an ensemble seems to be harder than attacking a single model, because an adversary must fool all models simultaneously. Some works on using a model ensemble to defend against adversarial examples~\citep{abbasi2017robustness,strauss2017ensemble,liu2018towards,pang2019improving,kariyappa2019improving} show promising results that they can indeed increase the required adversarial distortion for a successful attack and improve robustness. However, none of these works attempt to propose a \emph{provable} (or \emph{certifiable}) method to improve model robustness via an ensemble, so there is no guarantee that these methods work in all situations. 
For example, \citet{he2017adversarial} reported that attacking a specialist ensemble only increases the required adversarial distortion by as little as 6\% compared to a single model.

In this paper, we propose a new algorithm, \textbf{RobBoost}, that can provably enhance the robustness certificate of a deep model ensemble. First, we consider the setting where a set of pretrained robust models are given, and we aim to find the optimal weights for each base classifier that maximize provable robustness. We select the weight for each base model iteratively, to maximize a lower bound on the classification margin of the neural network ensemble classifier. Given a set of $T$ base models which are individually certifiable, we formulate a certified robustness bound for the model ensemble and show that solving the optimal ensemble leads to an optimization problem. We propose a coordinate descent based algorithm to iteratively solve the RobBoost objective. Second, we consider training each base classifier sequentially from scratch, in a setting more similar to traditional gradient boosting, 
where each model is sequentially trained to improve the overall robustness of the current ensemble.
Our experiments show that RobBoost can select a set of good base classifiers and weight them optimally, outperforming a naive average of all base models; on MNIST with $\ell_\infty$ perturbation of $\epsilon=0.3$, RobBoost reduces verified error from $36\%$ (averaging models) to $34\%$ using 12 certifiable base models trained individually.

\vspace{-5pt}
\section{Related Work}

\subsection{Neural network robustness verification}

The robustness of a neural network can be verified by analyzing the reachable range of an output neuron for \emph{all} possible inputs within a set (for example, a perturbed image with bounded $\ell_\infty$ norm), thus the margins of predictions between the top-1 class and other classes can be examined. Unfortunately, finding the exact reachable range is NP-complete~\citep{katz2017reluplex} and is equivalent to a mixed integer linear programming problem~\citep{tjeng2017evaluating,xiao2018training}. Therefore, many recent works in robustness verification develop computationally tractable ways to obtain outer bounds of reachable ranges. As robustness verification can be cast into a non-convex minimization problem~\citep{ehlers2017formal}, one approach is to resort to duality and give a lower bound of the solution~\citep{dvijotham2018dual,qin2018verification,dvijotham18verification}. Also, we can relax the primal optimization problem with linear constraints, and obtain a lower bound of the original problem using linear programming (LP) or the dual of LP~\citep{wong2018provable,wong2018scaling}. However, solving LPs for a relatively large network can still be quite slow~\citep{salman2019convex}. Fortunately, the relaxed LP problem can be solved greedily in the primal space~\citep{zhang2018crown,weng2018towards,wang2018efficient} or in the dual space~\citep{wong2018provable}. ``Abstract transformers''~\citep{singh2018fast,singh2019abstract,gehr2018ai,mirman2018differentiable,Singh2019robustness} propagate an abstraction of input regions layer by layer to eventually give us bounds on output neurons. See~\citep{salman2019convex,liu2019algorithms} for a comprehensive discussion on the connections between these algorithms.

Our work relies on the neural network outer bounds proposed in ~\citep{zhang2018crown,weng2018towards,wong2018provable}, which are the state-of-the-art methods to efficiently give an upper and a lower bound for an output neuron given an $\ell_p$-norm bounded input perturbation.  These bounds are essentially linear with respect to input perturbations, which is crucial for developing a tractable framework (see Proposition~\ref{prop:linear} for more details). Besides  these methods, local Lipschitz constants can also be used for efficiently giving a formal robustness guarantee~\citep{hein2017formal,zhang2018recurjac}; 
tighter bounds can be obtained by semi-definite programming (SDP) based methods~\citep{raghunathan2018certified,raghunathan2018semidefinite,dvijothamefficient2019}, although they scale much poorly to larger models.
\vspace{-5pt}
\subsection{Defending against adversarial examples}
We categorize existing defending techniques into two categories: \textit{certified} defenses that can provably increase the robustness of a model, and \textit{empirical} defenses that are mostly based on heuristics that have not been proven to improve robustness with a formal guarantee.
Empirical defenses, even sophisticated ones, can possibly be evaded using stronger or adaptive attacks~\citep{athalye2018obfuscated,carlini2017adversarial,carlini2017towards,carlini2017magnet}. We mostly focus on certified defenses in our paper, since our proposed method requires certified base models to create a certifiable ensemble.

Since many robustness verification methods give us a lower bound on the output margin between the ground-truth class and other classes under norm bounded input perturbation, optimizing a loss containing this bound obtained by a differentiable verification method will lead to maximizing this margin and improving \emph{verified error} on the training set.
For example, \citet{wong2018scaling,wong2018provable} propose to minimize the verified error through a cross-entropy loss surrogate using LP relaxation based verification bounds;  
\citet{wang2018mixtrain} optimize a similar verification bound; \citet{dvijotham2018training} proposes to learn the dual variables in dual relaxations using a learner network during training;
\citet{mirman2018differentiable} propose an differentiable version of abstract transformers and include the obtained bounds in loss function to provably increase robustness. \citet{raghunathan2018certified} propose to control the global Lipschitz constant of a 2-layer network to give a certified robustness bound, but it cannot be easily extended to multiple layers.



Previous \emph{certified} defenses mostly focus on improving the robustness of a single model. In~\citet{wong2018scaling}, multiple models are considered in a cascaded manner, where each subsequent model is trained using the examples that cannot be certified by previous models. In inference time, only the last cascaded model is considered. RobBoost works in a different scenario, where we consider how to combine certified base classifiers to a stronger one with better verified error. Empirical defenses using ensembles include~\citep{pang2019improving,kariyappa2019improving,abbasi2017robustness,strauss2017ensemble}, however they do not provide provable robustness guarantees.

\vspace{-5pt}
\section{The RobBoost Algorithm}

\newcommand{\tail}[2]{#1_{\overline{[#2]}}}
\newcommand{\abs}[1]{|#1|}
\newcommand{\tabs}[1]{\left|#1\right|}
\newcommand{\wh}{\widehat}
\newcommand{\wt}{\widetilde}
\newcommand{\ov}{\overline}
\newcommand{\eps}{\epsilon}
\newcommand{\N}{\mathcal{N}}
\newcommand{\R}{\mathbb{R}}
\newcommand{\volume}{\mathrm{volume}}
\newcommand{\RHS}{\mathrm{RHS}}
\newcommand{\LHS}{\mathrm{LHS}}
\newcommand{\bone}{\mathbf{1}}
\renewcommand{\i}{\mathbf{i}}
\newcommand{\norm}[1]{\left\lVert#1\right\rVert}
\renewcommand{\varepsilon}{\epsilon}
\renewcommand{\tilde}{\wt}
\renewcommand{\hat}{\wh}
\renewcommand{\R}{\mathbb{R}}
\renewcommand{\N}{\mathcal{N}}
\makeatletter
\newcommand*{\rom}[1]{\expandafter\@slowromancap\romannumeral #1@}
\makeatother

\newcommand{\x}{{\bm{x}}}
\newcommand{\deltax}{\Delta{\bm{x}}}
\newcommand{\xo}{{\bm{x}_0}}
\newcommand{\xn}{{\bm{x}_n}}
\newcommand{\xk}{\bm{x}_k}
\newcommand{\W}[1]{\mathbf{W}^{#1}}
\newcommand{\DD}[1]{\mathbf{D}^{#1}}
\newcommand{\Lam}[1]{\bm{\Lambda}^{#1}}
\newcommand{\upbias}[1]{\mathbf{T}^{#1}}
\newcommand{\lwbias}[1]{\mathbf{H}^{#1}}
\newcommand{\upbnd}[1]{\bm{u}^{#1}}
\newcommand{\lwbnd}[1]{\bm{l}^{#1}}
\newcommand{\z}{\bm{z}}
\newcommand{\y}{\bm{y}}
\newcommand{\bias}[1]{{b}^{#1}}
\newcommand{\setA}{\mathcal{A}}
\newcommand{\setIpos}[1]{\mathcal{I}^{+}_{#1}}
\newcommand{\setIneg}[1]{\mathcal{I}^{-}_{#1}}
\newcommand{\setIuns}[1]{\mathcal{I}_{#1}}
\newcommand{\set}[1]{\mathcal{#1}}
\newcommand{\Lipsloc}{L_{q,x_0}^j}

\newcommand{\gradl}[1]{\tilde{\bm{l}}^{#1}}
\newcommand{\gradu}[1]{\tilde{\bm{u}}^{#1}}
\newcommand{\gradC}[1]{\mathbf{C}^{#1}}
\newcommand{\gradL}[1]{\mathbf{L}^{#1}}
\newcommand{\gradU}[1]{\mathbf{U}^{#1}}
\newcommand{\gradLp}[1]{\mathbf{L'}^{#1}}
\newcommand{\gradUp}[1]{\mathbf{U'}^{#1}}
\newcommand{\Y}[1]{\mathbf{Y}^{#1}}
\newcommand{\Yp}[1]{\mathbf{Y'}^{#1}}

\newcommand{\Au}[1]{\mathbf{\Lambda}^{#1}}
\newcommand{\Al}[1]{\mathbf{\Omega}^{#1}}
\newcommand{\Du}[1]{\mathbf{\lambda}^{#1}}
\newcommand{\Dl}[1]{\mathbf{\omega}^{#1}}
\newcommand{\Ball}{{B}_{p}(\xk,\epsilon)}
\newcommand{\Ph}[1]{\Phi_{#1}}
\newcommand{\upslp}[2]{\mathbf{\alpha}^{#1}_{U,{#2}}}
\newcommand{\lwslp}[2]{\mathbf{\alpha}^{#1}_{L,{#2}}}
\newcommand{\upicp}[2]{\mathbf{\beta}^{#1}_{U,{#2}}}
\newcommand{\lwicp}[2]{\mathbf{\beta}^{#1}_{L,{#2}}}

\newcommand{\gradupbnd}[1]{\bm{u'}^{#1}}
\newcommand{\gradlwbnd}[1]{\bm{l'}^{#1}}
\newcommand{\M}{\mathbf{M}}
\newcommand{\setYp}[1]{\mathcal{T}^{+}_{#1}}
\newcommand{\setYn}[1]{\mathcal{T}^{-}_{#1}}
\newcommand{\setYo}[1]{\mathcal{T}_{#1}}
\newcommand{\A}{\mathbf{A}}
\newcommand{\B}{\mathbf{B}}
\newcommand{\bL}{\mathbf{L}}
\newcommand{\bLL}{\mathbf{\Lambda}}
\newcommand{\bU}{\mathbf{U}}
\newcommand{\bc}{\mathbf{c}}
\newcommand{\bcc}{\bm{\gamma}}
\newcommand{\bs}{\mathbf{s}}
\newcommand{\bd}{\mathbf{d}}
\newcommand{\bb}{\mathbf{b}}
\newcommand{\balp}{\bm{\alpha}}
\newcommand{\bara}{\bar{\bm{\alpha}}}
\newcommand{\bbta}{\bm{\beta}}
\newcommand{\gL}{\bm{g}_L}
\newcommand{\gc}{\bm{g}_c}

\newcommand{\overbar}[1]{\mkern 1.5mu\overline{\mkern-1.5mu#1\mkern-1.5mu}\mkern 1.5mu}
\newcommand{\hatW}[1]{\widehat{\mathbf{W}}^{#1}}
\newcommand{\hatA}{\widehat{\mathbf{A}}}
\newcommand{\barA}{\overbar{\mathbf{A}}}
\newcommand{\barL}{\overbar{\mathbf{L}}}
\newcommand{\tildeL}{\tilde{\mathbf{L}}}
\newcommand{\hatL}{\widehat{\mathbf{L}}}
\newcommand{\hatb}[1]{\hat{\mathbf{b}}^{#1}}
\newcommand{\hatc}{\hat{\mathbf{c}}}
\newcommand{\tildec}{\tilde{\mathbf{c}}}
\newcommand{\barc}{\bar{\mathbf{c}}}
\newcommand{\hatd}{\hat{\mathbf{d}}}

\renewcommand{\algorithmicrequire}{\textbf{Input:}}
\renewcommand{\algorithmicensure}{\textbf{Output:}}

\paragraph{Notations.} We define a $C$-class $H$-layer feed-forward classification neural network $f(\x) \in \R^C$ as
\begin{align*}
    f(\x)&=\W{(H)} h^{(H-1)} (\x) + \bb^{(H)} \\
    h^{(l)}(\x)&=\sigma^{(l)}(\W{(l)} h^{(l-1)}(\x)+\bb^{(l)}), \forall l
    \in\{1, \dots, H-1\}, 
\end{align*}
where  $h^{(0)} \coloneqq \x$ and layer $l$'s weight matrix is $\W{(l)} \in \R^{n_l \times n_{l-1}}$ and bias vector  is $\bb^{(l)} \in \R^{n_l}$. Input $\x \in \R^{d}$ and for convenience we define $n_0 = d$. $\sigma^{(l)}$ is a component-wise activation function. 
For convenience, we denote the row vector $\A_{(j,:)}$ as the $j$-th row of matrix $\A$, column vector $\A_{(:,i)}$ as the $i$-th column of matrix $\A$.
Additionally, 
We define the $\ell_p$ ball centered at $\x_0$ with radius $\epsilon$ as $B_p(\x_0, \epsilon) \coloneqq \{\x | \|\x - \x_0\|_p \leq \epsilon \}$, $1 \leq p \leq \infty$.  We use $[n]$ to denote the set $\{1, \cdots, n\}$. We denote a training example as a pair $(\xk, y_k)$, where $\xk \in \R^d$, $y_k \in [C]$ is the class label, $k \in [N]$ and $N$ is the total number of training examples.

\subsection{Linear outer bounds for neural networks}
\label{sec:linear_bounds}

We start with guaranteed linear upper and lower bounds for a single neural network $f(\x)$:

\begin{proposition}[Linear outer bounds of neural networks]
\label{prop:linear}
A neural network function $f(\x) \in \R^C$ can be linearly upper and lower bounded for all $\x = \xk + \deltax$, where  $\deltax \in B_p(0, \epsilon)$:
\begin{equation}
\label{eq:prop1}
\bL_{(j,:)} \deltax + \bc_j \leq [f(\x)]_j \leq \bU_{(j,:)} \deltax + \bd_j, \forall j \in [C]
\end{equation}
where $\bL, \bU \in \R^{C \times d}$and $\bc, \bd \in \R^{C}$ depend on $\xk$, $p$, $\epsilon$.
\end{proposition}

Proposition \ref{prop:linear} is a direct consequence of Theorem 3.2 in CROWN~\citep{zhang2018crown}, which gives the explicit form of $\bL, \bU, \bc, \bd$ as a function of neural network parameters. A similar outcome can be obtained from the neural network verification literature~\citep{wong2018provable,singh2018fast,weng2018towards,wang2018efficient}, but CROWN typically gives the tightest bound. We assume that $\bL, \bU, \bc, \bd$ are pre-computed for each example for a given $\epsilon$ using CROWN or other similar algorithms.

To verify if we can change the output of the network from the ground-truth class $i$ to another class $j$, we desire to obtain a lower bound on the margin $[f(\x)]_i - [f(\x)]_j$. For a training example $(\xk, y_k)$, we define the margin between the true class $i=y_k$ and other $C-1$ classes as a vector function $\hat{f}(\xk) \in \R^{C-1}$, where $[\hat{f}(\xk)]_j = [f(\xk)]_i - [f(\xk)]_j$, for $j < i$ and  $[\hat{f}(\xk)]_j = [f(\xk)]_i - [f(\xk)]_{j+1}$ for $j \geq i$, $j \in [C-1]$.
To obtain a lower bound of margin, we define the new network $\hat{f}(\x)$ with the same weights and biases as $f(\x)$, except that the last layer $H$ of  $\hat{f}(\x)$ is reformed as:
\begin{align}
\label{eq:w_diff}
\hatW{(H)}_{(j,:)} = \W{(H)}_{(i,:)} - \W{(H)}_{(s(j,i),:)}
\end{align}
\begin{align}
\label{eq:b_diff}
\hatb{(H)}_{j} = \bb^{(H)}_i - \bb^{(H)}_{s(j,i)}
\end{align}
\begin{equation*}
\text{where $j \in [C-1]$, } s(j,i) =
  \begin{cases}
  j \quad & j < i\\
  j+1 \quad & j \geq i
  \end{cases}   
\end{equation*}
$s(\cdot,i)$ is a $[C-1] \rightarrow [C]$ mapping for $j$ that skips the class $i$ (ground truth class). Then we can lower bound the margins $\hat{f}(\x)$ by Proposition~\ref{prop:linear}:
\[
\hatL_{(j,:)} \deltax + \hatc_j \leq [\hat{f}(\x)]_j, \forall j \in [C-1], 
\]
where $\hatL \in \R^{(C-1) \times d}$ and $\hatc \in \R^{C-1}$ implicitly depend on $(\xk, y_k)$, $\epsilon$ and $p$.
For \emph{every} $\x = \xk + \deltax$, $\deltax \in B_p(0, \epsilon)$, the following bound holds\footnote{As in~\cite{zhang2018crown} and many other works, we illustrate unbounded perturbation (not limited to $[0,1]$) to give the lower bound. Using bounded perturbation on MNIST typically improves verified error by 1-2\%.}:
\begin{equation}
\label{eq:lower_bound}
[\hat{f}(\x)]_j \geq \hatL_{(j,:)} \deltax + \hatc_j \geq - \eps \| \hatL_{(j,:)} \|_q + \hatc_j, 
\end{equation}
where $\|\cdot\|_q$ is the dual norm of $\|\cdot\|_p$. When $[\hat{f}(\x)]_j > 0$ for \emph{all} $\x \in B_p(\xk, \epsilon)$, we cannot change the classification of $\x$ from class $i$ to any other class, thus no adversarial examples exist. The $q$-norm of $\hatL_{(j,:)}$ plays an important rule in the model's robustness. We define the margin for target class $s(j,y_k)$ based on the guaranteed lower bound~\eqref{eq:lower_bound}:
\begin{equation}
\label{eq:margin}
    M(\xk,j) \coloneqq - \epsilon \| \hatL_{(j,:)} \|_q + \hatc_{j}
\end{equation}
With this lower bounds on margin, we can extend the definition of ordinary classification error and define \emph{verified error}, which is a provable upper bound on error under \emph{any} norm bounded attacks:

\begin{definition}[Robustness Certificate and Verified Error]
\label{def:certificate}
Given a perturbation radius $\epsilon$ where an input $\xk$ can be perturbed arbitrarily within $B_p(\x_k, \epsilon)$, verified error is the percentage of examples that do \emph{not} have a provable robustness certificate:
\begin{equation}
\label{eq:robust_err}
    \text{\textnormal{Verified Error}} \coloneqq 1 - \frac{\lvert \{ k | M(\xk,j) > 0 \text{\textnormal{ for all }} j \in [C - 1] \} \rvert}{N}
\end{equation}
\end{definition}

In our paper we also use a weaker definition of verified error (which can be bounded using a surrogate loss, which will be presented in Section~\ref{sec:ensemble}), where we consider each attack target individually:
\begin{equation}
\label{eq:targeted_robust_err}
    \text{Targeted Verified Error} \coloneqq 1 - \frac{\lvert \{ (k, j) | M(\xk,j) > 0 \} \rvert}{N \times (C-1)}.
\end{equation}

\subsection{Linear outer bounds for an ensemble}
\label{sec:linear_ensemble}

We denote $M_t(\xk,j)$ as the margin for a base model $f_t(\x)$. We first note that the lower bound on margin $M_t(\xk,j)$ is unnormalized and not scale-invariant. Enlarging the output of $f_t(\x)$ by a constant factor does not affect the robustness of $f_t(\x)$, but the margin will also be scaled correspondingly. Directly maximizing this unnormalized margin does not lead to better robustness. Intuitively, the most important factor of the robustness of model $f_t(\x)$ is the available budget (reflected as the term $\hatc_{t,j}$) divided by the sensitivity with respect to the input (reflected as $\| \hatL_{t,(j,:)} \|_q$), rather than the absolute value of the margin.
When we use this lower bound on margins as a surrogate to compare the robustness across different models, their margins should be within  a similar range to make this comparison meaningful.
To take this into account, we can normalize $M_t(\xk,j)$ by dividing it with a normalizing factor $Z_t \coloneqq \frac{1}{N(C-1)} \sum_{k = i}^N \sum_{j \in [C-1]} \hatc^k_{t,j}$, the average $\hatc$ over all examples and classes. The normalized lower bound on margin is defined as:
\begin{align}
\label{eq:margin_normalized}
    \tilde{M}_t(\xk,j) \coloneqq& - \epsilon \| \tildeL_{t,(j,:)} \|_q + \tildec_{t,j},\\ \qquad \text{where }
    \tildeL_{t,(j,:)} \coloneqq& \frac{1}{Z_t} \hatL_{t,(j,:)}, \quad \tildec_{t,j} \coloneqq \frac{1}{Z_t} \hatc_{t,j}. \nonumber
\end{align}
This is equivalent to applying a constant factor $\frac{1}{Z_t}$. The normalized model, denoted as $\tilde{f}_t(\x) \coloneqq \frac{1}{Z_t} f_t(\x)$, will be used for our ensemble. We can use other normalizing schemes as long as they roughly keep each model's margin in a similar magnitude to ease the optimization.

We define the model ensemble of a set of $T$ neural networks as $F(\x) = \sum_{t=1}^T \alpha_t \tilde{f}_t(\x)$, where $0 \leq \alpha_t \leq 1$, $\sum_{t=1}^T \alpha_t = 1$. $\alpha_t$ is the coefficient for classifier $\tilde{f}_t(x)$. In our setting, each neural network classifier has been trained with certificates (Def.~\ref{def:certificate}), and we want to further enhance them using carefully selected ensemble weights. Because they are linearly combined, we can give a linear upper and lower bound for $F(\x)$, by linearly combining the upper and lower bounds given by Eq.~\eqref{eq:prop1}.
Again, for a training example $(\xk, y_k)$, the lower bound on the margin $[\hat{F}(\x)]_j$ of $F(\x)$ for the class $j \neq y_k$ inside $\Ball$ is:
\begin{equation}
\label{eq:ens_bnd}
\left ( \sum\nolimits_{t=1}^T \alpha_t \tildeL_{t,(j,:)} \right ) \deltax + \sum\nolimits_{t=1}^T \alpha_t \tildec_{t,j} \leq [\hat{F}(\x)]_j
\end{equation}
Analogous to Eq.~\eqref{eq:lower_bound}, these bounds are \textit{guaranteed} for any $\x = \xk + \deltax$, $\deltax \in B_p(0, \epsilon)$. $\tildeL_t \in \R^{(C-1) \times d}, \tildec_t \in \R^{C-1}$ are computed for model $\tilde{f}_t(\x)$ using Proposition~\ref{prop:linear} with a similar network transformation as in Eq.~\eqref{eq:w_diff} and~\eqref{eq:b_diff} for each $\tilde{f}_t(\x)$. Note that each $f_t(\x)$ can have \emph{completely different internal structure} (number of layers, number of neurons, architecture, etc) and be trained using different schemes, but their corresponding $\tildeL_t$ and $\tildec_t$ have the same dimension.
For the ensemble, the normalized lower bounds of margin is the following:
\begin{equation}
\label{eq:margin_ensemble}
    M(\xk,j) \coloneqq - \epsilon \| \sum\nolimits_{t=1}^T \alpha_t \tildeL_{t,(j,:)} \|_q + \sum\nolimits_{t=1}^T \alpha_t \tildec_{t,j}
\end{equation}

\iftoggle{LONG_VERSION}{
\begin{figure}[b]
  \centering
  \includegraphics[trim={2cm 2cm 2cm 2cm},width=0.7\linewidth]{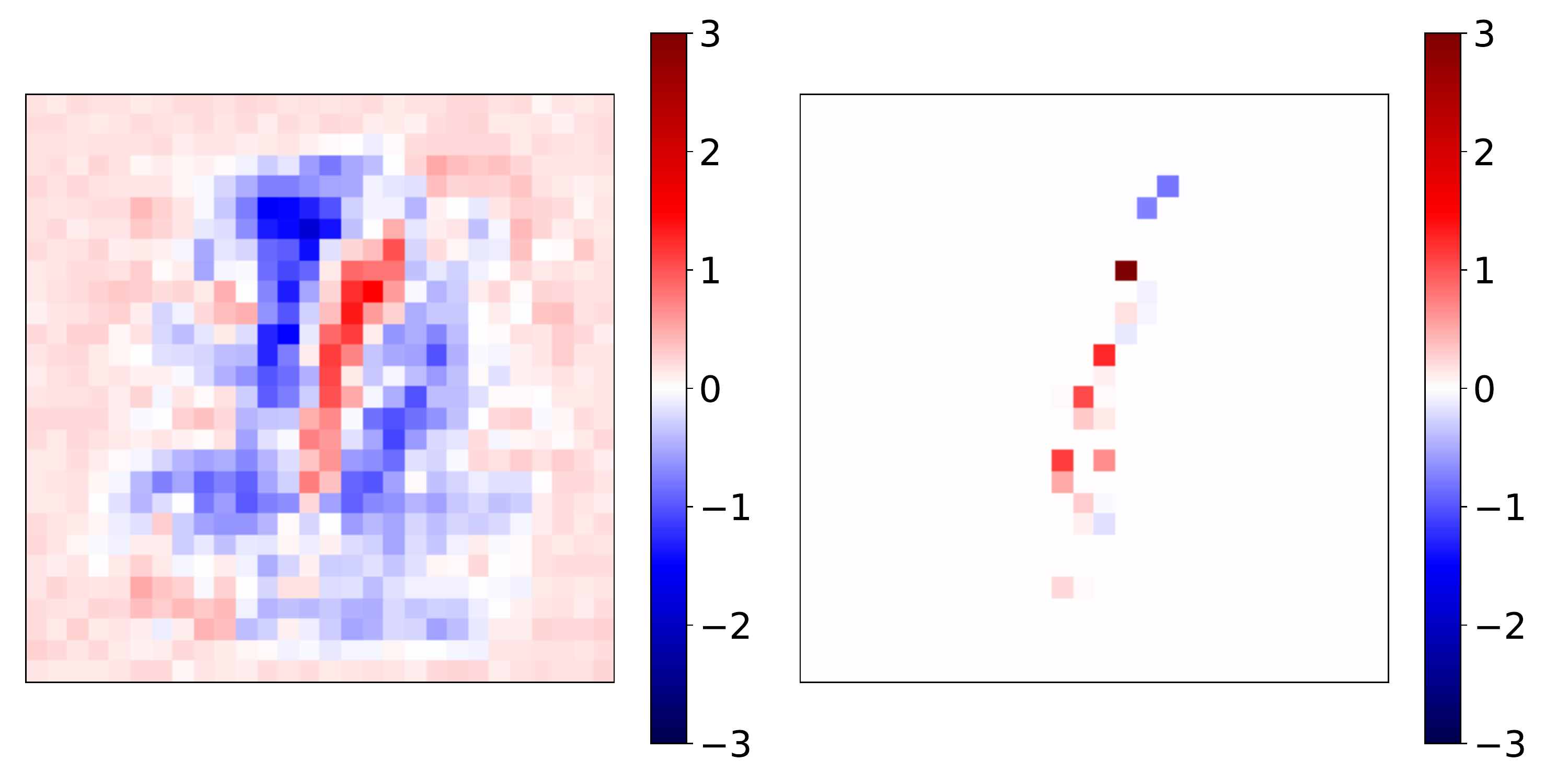}
  \caption{Left: visualization of matrix $\tildeL$ for a naturally trained MNIST model. Right: visualization of matrix $\tildeL$ for an adversarially trained MNIST model. Both models are 4-layer MLPs with 1024 neurons per layer. We use an image of digit ``1'' ($i=1$) as the input, and set attack target to ``2'' ($s(j,i)=2$), $\epsilon=0.05$.}
  \label{fig:adv_vs_nat}
\end{figure}
}

{\bf Main Idea: Why an ensemble can improve robustness?} To enhance robustness via an ensemble, we want to find an optimal $\bm{\alpha}$, such that $M(\xk,j)$ is maximized for all training examples $k \in [N]$ and target classes $s(j,y_k), j \in [C-1]$. A model $\tilde{f}_t(\x)$ lacking of robustness often has large $\| \tildeL_{t,(j,:)} \|_q$; by combining $\tildeL_{t,(j,:)}$ of different models with optimal weights, we hope that some noises in $\tildeL_{t,(j,:)}$ can be canceled out and $\sum\nolimits_{t=1}^T \alpha_t \tildeL_{t,(j,:)}$  has a smaller $q$-norm than $\tildeL_{t,(j,:)}$. When there are very limited number of models for selection, a naive ensemble (with all $\alpha_t = \frac{1}{T}$) cannot guarantee to achieve this goal. Instead, \textbf{RobBoost} optimally select $\alpha_t$ based on maximizing a surrogate loss on normalized lower bounds of margins.

\iftoggle{LONG_VERSION}{
In Figure~\ref{fig:adv_vs_nat}, we plot matrix $\tildeL$ for a naturally trained and a robust model (adversarially trained using~\citep{madry2018towards}) to show the intuition behind RobBoost. Strikingly, $\tildeL$ has a quite interpretable pattern, especially on the adversarially trained model -- $\tildeL$ is surprisingly sparse and the model output is only sensitive to changes on the pixels of the digit ``1''; on the other hand, the naturally trained model has a lot of random noise around the ``1'' in the center, and is sensitive to many irrelevant background pixel changes. Our aim is thus to make the matrix $\tildeL$ less ``noisy'' through an careful ensemble: $\tildeL = \sum\nolimits_{t=1}^T \alpha_t \tildeL_t$.
}{}

\subsection{The RobBoost Loss Function}
\label{sec:ensemble}

We first define the following RobBoost loss to maximize model's lower bound of margin via ensemble, across all examples and classes:
\begin{equation}
\label{eq:main_obj}
\begin{split}
    \text{minimize} \enskip g(\balp) =& \sum_{k \in [N]} \sum_{j \in [C-1]} \max(1-M(\xk, j), 0) \qquad \\
    \text{s.t. } 0 \leq \balp& \leq 1, \quad \sum\nolimits_{t=1}^T \alpha_i = 1
\end{split}
\end{equation}
where $\balp \in \R^T$ is the vector of weights for each model $\tilde{f}_t(\x)$, and $M(\xk,j)$ is defined as in Eq.~\eqref{eq:margin_ensemble}. This is a hinge-style surrogate loss of the 0-1 error defined in~\eqref{eq:targeted_robust_err} that encourages large margin. We aim to decrease~\eqref{eq:targeted_robust_err} by using an optimally weighted ensemble. 

In this paper, we focus on the most common threat model where $\ell_\infty$ adversarial distortion ($p=\infty, q = 1$) is applied ($\ell_\infty$-RobBoost). In this case,  Eq.~\eqref{eq:margin_ensemble} is a summation of absolute values:
\begin{align}
\label{eq:sum_abs}
    M(\xk,j) \coloneqq - \epsilon \sum_{i=1}^d | \sum_{t=1}^T \alpha_t \tildeL^k_{t,(j,i)} | + \sum_{t=1}^T \alpha_t \tildec^k_{t,j}
\end{align}
Here we explicitly write out the dependency of $k$ in $\tildeL$ and $\tildec$.
When $j, k$ is fixed, we can define $N \times (C-1)$ new matrices $\barA \in \R^{d \times T}$ with reordered indices and absorb $\epsilon$ into it: $\barA^{j,k}_{(i,t)} \coloneqq \epsilon \tildeL^{k}_{t,(j,i)}$. And similarly $\barc_{t}^{j,k} \coloneqq \tildec_{(t,j)}^k$ such that Eq.~\eqref{eq:sum_abs} can be rewritten as:
\begin{equation*}
     M(\xk,j) = - \| \barA^{j,k} \balp \|_1 + \barc^{j,k{\top}} \balp
\end{equation*}
Note that for each example $k$ and each target class $j$ we have an $\barA$ and a $\barc$.
Then Eq.~\eqref{eq:main_obj} becomes:
\begin{equation}
\label{eq:rearranged_obj}
\begin{split}
    & \text{minimize} \enskip g(\balp) = \\ 
    \sum_{k \in [N]} \sum_{j \in [C-1]}& \max(\| \barA^{j,k} \balp \|_1 - \barc^{j,k \top} \balp + 1, 0) \\
    \text{s.t. }& 0 \leq \balp \leq 1, \quad \sum\nolimits_{t=1}^T \alpha_i = 1
\end{split}
\end{equation}
Since the objective is non-smooth, piece-wise linear and low-dimensional, we propose to use coordinate descent to efficiently solve this minimization problem.

\paragraph{Solving $\ell_\infty$-RobBoost using coordinate-descent.} In coordinate descent, we aim to solve the following one variable optimization problem with a randomly selected coordinate $t$:
\begin{equation}
\label{eq:coord_opt}
\begin{split}
    \alpha_t^* &= \argmin_{0 \leq \alpha_t \leq 1} g(\alpha_t), \text{ where} \\
    g(\alpha_t) &= \sum_{k \in [N]} \sum_{j \in [C-1]} \max(\| \barA^{j,k} \balp \|_1 - \barc^{j,k\top} \balp + 1, 0) \\
    &= \sum_{k \in [N]} \sum_{j \in [C-1]} \max(\|  \alpha_t \barA^{j,k}_{(:,t)} + \barA^{j,k} (\balp - \alpha_t \bm{e}_t)  \|_1 -\\ & \barc^{j,k\top} (\balp - \alpha_t \bm{e}_t) - \barc^{j,k} \alpha_t + 1, 0) \\
    &\coloneqq \sum_{k \in [N]} \sum_{j \in [C-1]} \max(\|  \alpha_t \barA^{j,k}_{(:,t)} + \bm{v}^{j,k}  \|_1 + \\ & u^{j,k} - \barc^{j,k \top} \alpha_t + 1, 0)\\
    &\text{s.t.} \sum\nolimits_{t=1}^T \alpha_i = 1
\end{split}
\end{equation}
where $\bm{v}^{j,k} = \barA^{j,k} (\balp - \alpha_t \bm{e}_t)$ and $u^{j,k} = \barc^{j,k\top} (\balp - \alpha_t \bm{e}_t)$ are constants with respect to $\alpha_t$. While the box constraint $0 \leq \alpha_t \leq 1$ can be easily handled by coordinate-descent, the challenge is the constraint $\sum_{t=1}^T \alpha_i = 1$, which needs to be directly enforced during the coordinate descent procedure. Suppose we have maintained $\sum\nolimits_{t=1}^T \alpha_i = 1$ before update, when updating variable $\alpha_t$, we enforce this constraint by scaling all other $\alpha_s, s \neq t$ by a factor of $\frac{1-\alpha_t}{\sum_{s \neq t}\alpha_s}$ such that $\alpha_t + \sum_{s \neq t}\alpha_s \equiv 1$. In other words, we redefine $g(\alpha_t)$ as:
\begin{equation}
\begin{split}
&g(\alpha_t) = \sum_{k \in [N]} \sum_{j \in [C-1]} \max(\|  \alpha_t \barA^{j,k}_{(:,t)} + \frac{1-\alpha_t}{\sum_{s \neq t}\alpha_s}\bm{v}^{j,k}  \|_1 \\ &+ \frac{1-\alpha_t}{\sum_{s \neq t}\alpha_s} u^{j,k} - \barc^{j,k \top} \alpha_t + 1, 0) \\
&\coloneqq \sum_{k \in [N]} \sum_{j \in [C-1]} \max(\|  \alpha_t \bm{\omega}^{j,k} + \bm{\nu}^{j,k}  \|_1 + \gamma^{j,k} \alpha_t + \mu^{j,k} + 1, 0) \\
&\coloneqq \sum_{k \in [N]} \sum_{j \in [C-1]} \max(g^{j,k}(\alpha_t) ,0) 
\end{split}
\label{eq:coord_rescale}
\end{equation}
where $\bm{\omega}^{j,k} \coloneqq \barA^{j,k}_{(:,t)} - \frac{1}{\sum_{s \neq t}\alpha_s} \bm{v}^{j,k}$, $\bm{\nu}^{j,k}=\frac{1}{\sum_{s \neq t}\alpha_s}\bm{v}^{j,k}$ and $\mu^{j,k}=\frac{u^{j,k}}{\sum_{s \neq t}\alpha_s} + 1$, $\gamma^{j,k}=-\barc^{j,k \top} -\frac{1}{\sum_{s \neq t}\alpha_s}$. Now the summation constraint has been removed, and for any $0 \leq \alpha_t \leq 1$ we guarantee $\sum\nolimits_{t=1}^T \alpha_i = 1$.
Each term $g^{j,k}(\alpha_t)$ is a bounded one-dimensional piece-wise linear function within domain $0 \leq \alpha_t \leq 1$. The term $\| \alpha_t \bm{\omega}^{j,k} + \bm{\nu}^{j,k} \|_1$ contains $d$ linear terms inside absolute value so there are at most $d$ pieces. The minima of this term must be on the end of one piece, or at the boundary 0 or 1. Solving for $\alpha_t$ in $d$ equations $\alpha_t \bm{\omega}^{j,k}_l + \bm{\nu}^{j,k}_l = 0, l \in [d]$ gives us the locations of the end points of these pieces in $O(d)$ time, denoted as $r^{j,k}_1, \cdots, r^{j,k}_{d}$. We consider the worst case where all $d$ points lie in $(0,1)$, and for convenience we denote $r^{j,k}_0 = 0$, $r^{j,k}_{d+1} = 1$.
Now the challenge remains to efficiently evaluate $g^{j,k}(\alpha_t)$ at these $d$ endpoint and two boundaries, each in $O(1)$ time. We first sort $r^{j,k}_1, \cdots, r^{j,k}_{d}$ in ascending order as $r^{j,k}_{\pi(1)}, \cdots, r^{j,k}_{\pi(d)}$, where $\pi$ is the permutation of sorting and we additionally define $\pi(0):=0$. Then, we start with $r^{j,k}_0 = 0$, and check the sign of $z^{j,k}_l(\alpha_t) \coloneqq \alpha_t \bm{\omega}^{j,k}_l + \bm{\nu}^{j,k}_l$ at $\alpha_t = r^{j,k}_0$ for each $l \in [d]$. We define two sets indicating the sign of $z^{j,k}_l$:
\begin{align}
\label{eq:coord_j0+}
J^{j,k}_+ &= \{l | z^{j,k}_l(0) > 0, \text{ or } z^{j,k}_l(0)=0, \bm{\omega}^{j,k}_l > 0 \} \\
\label{eq:coord_j0-}
J^{j,k}_- &= \{l | z^{j,k}_l(0) < 0, \text{ or } z^{j,k}_l(0)=0, \bm{\omega}^{j,k}_l < 0 \} 
\end{align}

They can be formed in $O(d)$ time. Then we define the effective slope and intercept at the point $r_0$ as:

\begin{align}
\label{eq:coord_a0}
a^{j,k}_0 &= \sum_{l \in J^{j,k}_+} \bm{\omega}^{j,k}_l - \sum_{l \in J^{j,k}_-} \bm{\omega}^{j,k}_l + \gamma^{j,k}\\
\label{eq:coord_b0}
b^{j,k}_0 &= \sum_{l \in J^{j,k}_+} \bm{\nu}^{j,k}_l - \sum_{l \in J^{j,k}_-} \bm{\nu}^{j,k}_l + \mu^{j,k}
\end{align}

and then we can evaluate $g^{j,k}(r^{j,k}_0) = a^{j,k}_0 \times r^{j,k}_0 + b^{j,k}_0$. For $i = \{1, \cdots, N^\prime\}$, we evaluate $g^{j,k}(r^{j,k}_{\pi(i)})$ one by one. Assuming we already obtained $a^{j,k}_{m}$ and $b^{j,k}_{m}$ ($m < d$) and evaluated $g^{j,k}(r^{j,k}_{\pi(m)}) = a^{j,k}_{m} r^{j,k}_{\pi(m)} + b^{j,k}_{m}$. We can then recursively define $a^{j,k}_{m+1}$ and $b^{j,k}_{m+1}$ as:
\begin{align}
\begin{split}
    a^{j,k}_{m+1} = a^{j,k}_{m}  &+ I(\pi(m+1) \in J^{j,k}_-) \cdot 2 \bm{\omega}^{j,k}_{\pi(m+1)} \\
    &- I(\pi(m+1) \in J^{j,k}_+) \cdot 2 \bm{\omega}^{j,k}_{\pi(m+1)} \label{eq:update_a} 
\end{split}
\\
\begin{split}
    b^{j,k}_{m+1} = b^{j,k}_{m} &+ I(\pi(m+1) \in J^{j,k}_-) \cdot 2 \bm{\nu}^{j,k}_{\pi(m+1)} \\
    &- I(\pi(m+1) \in J^{j,k}_+) \cdot 2 \bm{\nu}^{j,k}_{\pi(m+1)} \label{eq:update_b}
\end{split}
\end{align}
$I(\cdot)$ is an indicator function. We keep maintaining the slope and intercept for the next linear piece, when the sign of term $\pi(m+1)$ just changed. This update only takes $O(1)$ time. For $m = d+1$, we reached the boundary 1 and evaluate $g^{j,k}(1) = a^{j,k}_{d} \times 1 + b^{j,k}_{d}$.

For the final sum of surrogate losses, we merge sort all $r^{j,k}_i$ for all $(j,k)$ terms into a new vector $r$ with $N \times (C-1) \times d$ elements, also maintain a mapping $p(i) \rightarrow (j,k)$ which maps an element $r_i$ into the summation term $(j,k)$ it comes from. Our final algorithm evaluates the objective function on all $r_i$ and using the maintained effective slope $a^{j,k}$ and intercept $b^{j,k}$ for each term in summation, and update $a^{p(i)}$ and $b^{p(i)}$ using~\eqref{eq:update_a} and~\eqref{eq:update_b}. Additionally, we need to consider up to $NC$ additional linear pieces introduced by the $\max$ function. We list the full algorithm in appendix in Algorithm~\ref{alg:coord_hinge}. 
In each iteration of coordinate descent, we randomly choose a coordinate $t$, and obtain the best value $\alpha_t^*$~\eqref{eq:coord_opt} to minimize the loss and set $\alpha_t \leftarrow \alpha_t^*$. Then we choose another coordinate and repeat. We observe that 2 to 3 epochs (each epoch visits all coordinate once) are sufficient to find a good solution.

\subsection{RobBoost in Gradient Boosting}

Gradient Boosting builds a strong model $F(\x)$ by iteratively training and combining weak models:
\[
F_{m+1}(\x) = F_m(\x) + f_{m+1}(\x)
\]
where $F_m(\x)=\sum_{t=1}^m f_{t}(\x)$ is kept unchanged and we train $f_{m+1}(\x)$ to reduce a certain loss function on $F_{m+1}(\x)$. Unlike the setting we discussed in Section~\ref{sec:ensemble} where all base models are given and fixed, here we are allowed to update the model parameters of the last base model, with previous models frozen. Similar to Eq.~\eqref{eq:margin_ensemble}, we can write the margin for example $\xk$ target class $j$ for the setting of gradient boosting:
\begin{equation}
\label{eq:margin_gb}
\begin{split}
    M(\xk,j) \coloneqq& - \epsilon \| \sum\nolimits_{t=1}^m \hatL_t + \hatL_{m+1} \|_q\\ 
    &+ \sum\nolimits_{t=1}^m \hatc_t + \hatc_{m+1}
\end{split}        
\end{equation}

Suppose we have some surrogate loss function $\ell$, we define the following loss function:
\begin{equation}
\label{eq:gb_loss}
\begin{split}
L \coloneqq& \sum_{k \in [N]} \sum_{j \in [C-1]} \ell(M(\xk,j))\\
=& \sum_{k \in [N]} \sum_{j \in [C-1]} \ell(- \epsilon \| \sum\nolimits_{t=1}^m \hatL_t + \hatL_{m+1} \|_q \\&+ \sum\nolimits_{t=1}^m \hatc_t + \hatc_{m+1})
\end{split}
\end{equation}
Note that all $\hatL_t, \hatc_t$ for $t=\{1, \cdots, m\}$ are precomputed and can be treated as constants, and $\hatL_{m+1}, \hatc_{m+1}$ are functions of the neural network parameters $\W{}_{m+1}, \bb{}_{m+1}$ of model $f_{m+1}$ (due to  Prop.~\ref{prop:linear},  see~\cite{wong2018provable,zhang2018crown} for the explicit form). We can thus take the gradient $\frac{\partial L}{\partial \W{}_{m+1}}$, $\frac{\partial L}{\partial \bb{}_{m+1}}$ and use typical gradient-based optimization tools to update model $f_{m+1}(x)$ and reduce the loss $L$. Compared to the setting in the previous section where each base model is trained independently and then fixed, in~\eqref{eq:gb_loss}, model $f_{m+1}(x)$ knows the ``weakness'' of the ensemble of all previous models, and attempts to ``fix'' it, offering more flexibility.

\vspace{-5pt}
\section{Experiments}
\vspace{-5pt}
\setlength{\textfloatsep}{5pt}
\paragraph{Overview and Setup.}
We evaluate the effectiveness of RobBoost, by using it to find the best ensemble in a relatively large pool of models on MNIST and CIFAR-10 datasets. Since we focus on improving certified robustness, our main metric to evaluate model robustness is \emph{verified error} on test set, as defined in~\eqref{eq:robust_err}; this is a provable upper bound of PGD attack error and has been  used as the standard way to evaluate certified defense methods~\citep{wong2018provable,wong2018scaling}. Since there is no existing work on boosting provable robustness, our baseline is the naive ensemble, where each model is equally weighted. Because our purpose is to show how optimally RobBoost weights each base model, we do not focus on tuning each base model to achieve state-of-the-art results on each dataset. We use small base models, where all MNIST models sum to 9.1 MB and all CIFAR models sum to 10.0 MB. We precompute $\hatL$ for all training examples for each model. This precomputation takes similar time as 1 epoch of robust training~\citep{wong2018provable}, since they need to compute the same bounds every epoch for training a single robust model, and they typically need 100 to 200 epochs for training. The time of our experiments is dominated by training each base models (hours to days each) rather than precomputing these matrices and solving the ensemble objective (1-2 hours).


\paragraph{Data Elimination.} We first remove all data points that have positive margins on \emph{all} base models, as they will remain robust regardless of any positive weights. Similarly, we remove all data points that have negative margins on \emph{all} base models, as the ensemble is not capable to improve robustness for them. These pre-processing steps allow us to focus on the data points whose robustness can be potentially enhanced, and also reduce the effective training data size. In Table~\ref{tab:ex_eli}, we report the percentage of examples eliminated in this preprocessing step. For all models, a large portion of examples (ranging from 70\% to 97\%) can be excluded from the optimization step, greatly improving the efficiency of RobBoost.

\begin{table}[tbp]    
    \caption{Examples eliminated in preprocessing}
    \label{tab:ex_eli}
        \centering
    \scalebox{0.75}{
    \begin{tabular}{|c|c|c|c|}
    \hline
         \multirow{2}{*}{Dataset}& \multirow{2}{*}{$\epsilon$}  & Eliminated because  & Eliminated because NOT\\
          &  & robust to \textit{all} models  & robust to \textit{all} models \\
           \hline
           \multirow{3}{*}{MNIST} & 0.1 & 97.67\%& 0.07\% \\\cline{2-4}
           & 0.2 & 91.85\%&0.42\% \\ \cline{2-4}
           & 0.3 & 69.69\%&2.84\%\\ \cline{2-4}
           \hline
           \multirow{2}{*}{CIFAR} & 2/255 &65.60\%&3.97\% \\\cline{2-4}
           & 8/255 &56.23\%&14.11\% \\ 
           \hline
    \end{tabular}
    }
\end{table}

\begin{table*}[ht]
\caption{Comparison of verified error, clean error and PGD error of RobBoost and naive ensembles. As a comparison to single model defense, for MNIST $\epsilon=0.3$, the best model in~\cite{wong2018scaling} achieves 43.10\% verified error and 14.87\% clean error; for CIFAR $\varepsilon = \frac{8}{255}$, the ResNet (best) model in~\cite{wong2018scaling} achieves 78.22\% verified error and  71.33\% clean error. All our CIFAR models combined have a smaller size (10 MB) than the ResNet model in~\cite{wong2018scaling} (\textbf{40 MB}), and the RobBoost ensemble outperforms the single largest model in~\cite{wong2018scaling} in both clean and verified error.
}
\label{tb:robust_result}
\centering
\begin{tabular}{|c|c|c|c|c|c|c|c|}
\hline
\multirow{2}{*}{Model} & \multirow{2}{*}{$\epsilon$} & \multirow{2}{*}{Ensemble} & \multicolumn{2}{c|}{Verified Error}& Clean Error      & PGD Error        & Ensemble   \\ \cline{4-7} 
                       &                          &                           & Train            & Test             & Test             & Test             & Model Size \\ \hline
\multirow{6}{*}{MNIST} & \multirow{2}{*}{0.1}     & Naive                     & 3.55\%           & 4.56\%           & 0.39\%           & 1.56\%           &9.1M \\ \cline{3-8} 
                       &                          & RobBoost                  & \textbf{2.65\%}  & \textbf{4.47\%}  & 0.39\%           & \textbf{1.17\%}  &\bf 5.3M \\ \cline{2-8} 
                       & \multirow{2}{*}{0.2}     & Naive                     & 13.82\%          & 13.66\%          & 2.34\%           & 3.91\%           &9.1M  \\ \cline{3-8} 
                       &                          & RobBoost                  & \textbf{12.41\%} & \textbf{12.58\%} & \textbf{1.95\%}  & \textbf{3.12\%}  &\bf 7.4M \\ \cline{2-8} 
                       & \multirow{2}{*}{0.3}     & Naive                     & 39.22\%          & 38.60\%          & 11.33\%          & 19.92\%          &9.1M  \\ \cline{3-8} 
                       &                          & RobBoost                  & \textbf{37.42\%} & \textbf{36.61\%} & \textbf{9.77\%}  & \textbf{19.53\%} &\bf 5.7M \\ \hline
\multirow{4}{*}{CIFAR} & \multirow{2}{*}{2/255}   & Naive                     & 50.02\%          & 51.68\%          & 38.28\%          & 45.31\%          &10.0M \\ \cline{3-8} 
                       &                          & RobBoost                  & \textbf{47.27\%} & \textbf{49.51\%} & \textbf{34.77\%} & \textbf{41.80\%} & \bf 6.1M \\ \cline{2-8} 
                       & \multirow{2}{*}{8/255}   & Naive                     & 74.95\%          & 74.50\%          & 62.89\%          & 71.09\%          &10.0M  \\ \cline{3-8} 
                       &                          & RobBoost                  & \textbf{74.56\%} & \textbf{74.27\%} & \textbf{60.94\%} & \textbf{69.14\%} & \bf 5.6M\\ \hline
\end{tabular}
\end{table*}

\vspace{-5pt}
\paragraph{Ensemble of robustly trained models with different architectures.} We train 12 MNIST models and 11 CIFAR models with a variety of architectures. The models are trained using convex adversarial polytope~\citep{wong2018scaling}, a certified defense method, under different $\ell_\infty$ norm perturbations ($\epsilon = \{0.1, 0.2, 0.3\}$ for MNIST and $\epsilon = \{\frac{2}{255}, \frac{8}{255}\}$ for CIFAR). Even the largest CIFAR model used is much smaller than the best ResNet model in~\citet{wong2018scaling} (40 MB), as we desire to use an ensemble of small models to obtain better robustness than a single large model. For CIFAR, the smallest model is around 0.1 MB and the largest is 3 MB\iftoggle{LONG_VERSION}{; see more details on model structure in Appendix~\ref{sec:models}}{}. We list results in Table~\ref{tb:robust_result}. We can observe that RobBoost consistently outperforms naive averaging ensemble in both verified and clean error. Also the ensemble model we created performs better in all metrics than the single large model reported in literature~\citep{wong2018scaling}.

\iftoggle{LONG_VERSION}{
In Figure \ref{fig:margin_distribution}, we plot the distributions of lower bounds of margins $M(\xk, j)$ for two models. A value less than 0 indicates that an example cannot be certified. Compared to the naive ensemble, we can clearly observe that the distributions of margins for RobBoost ensembles have more mass on the positive side, reflecting the improvements on certified robustness.
}{}

\begin{figure}[htbp]
     \centering
     \begin{subfigure}[t]{0.35\textwidth}
         \centering
         \includegraphics[width=\textwidth]{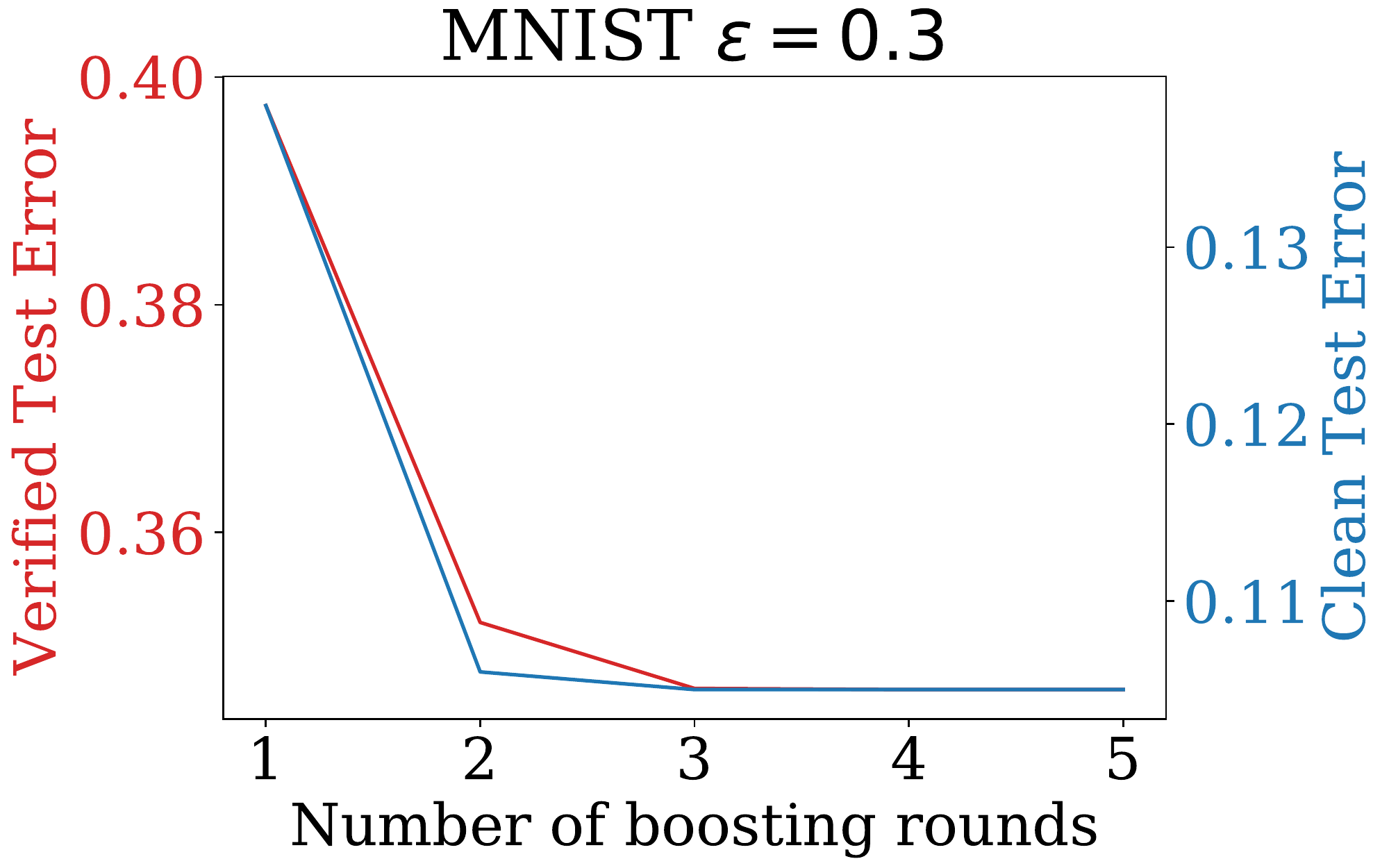}
     \end{subfigure}
    ~
     \begin{subfigure}[t]{0.35\textwidth}
         \centering
         \includegraphics[width=\textwidth]{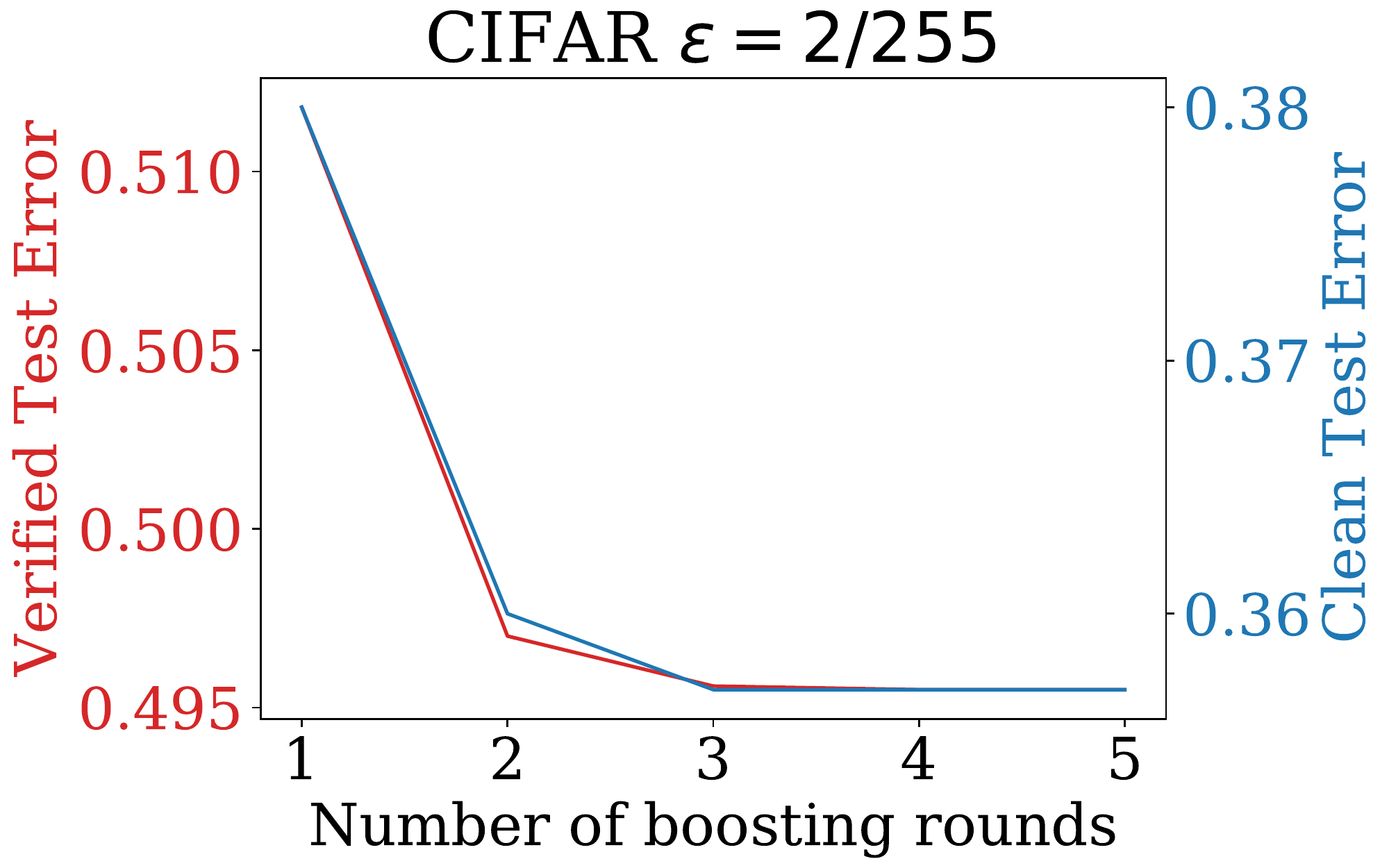}
     \end{subfigure}
  \caption{RobBoost in gradient boosting setting. With 3-5 base models RobBoost significantly reduces both verified and clean errors on both CIFAR and MNIST datasets compared to a single base model.}
  \label{fig:grad_boost}
\end{figure}

\paragraph{Ensemble of the same model architecture with feature subsampled data.} A common practice in building traditional ensemble models like random forest is to use feature subsampling, i.e., each base model only uses a subset of the features to train. In this experiment, we use only 1 model structure but randomly sample 80\% pixels to train the model (a recent work~\citep{hosseini2019dropping} presented a similar idea, but their method is not a certified defense). We train 5 feature sub-sampled models for MNIST with $\epsilon=0.3$ and CIFAR with $\epsilon=\frac{2}{255}$. In Table \ref{tab:subsample}, we list the verified error and clean error of each model for both the naive ensemble and RobBoost ensemble. Since there are only 5 base models, RobBoost ensembles provide a small but consistent performance advantage in verified error.
\begin{table*}[ht]    
    \caption{Naive and RobBoost ensembles on 5 small models with the same architecture but trained on different feature subsamples. Since they have the same architecture, their robustness are similar and the optimal weight for each model is close to $\frac{1}{5}$; in this case the naive baseline is close to RobBoost as it is a special case of RobBoost.}
    \label{tab:subsample}
    \centering
    \begin{tabular}{|c|c|c|c|c|c|c|c|c|c|}
    \hline
         \multirow{2}{*}{Dataset}& \multirow{2}{*}{$\epsilon$} &\multirow{2}{*}{Test Error} &  \multicolumn{5}{c|}{Base Models No.} &\multirow{2}{*}{Naive} & \multirow{2}{*}{RobBoost} \\ \cline{4-8} 
           & & & 1 & 2 & 3 & 4 & 5 &  &  \\
           \hline
           \multirow{2}{*}{MNIST} & \multirow{2}{*}{0.3} & Clean&12.96\% & 13.99\% & 23.88\% & 12.77\% & 18.50\% &  13.34\%& {\bf 12.51\%}\\ \cline{3-10}
            & & Verified&42.98\% & 45.3\% & 51.62\% & 45.13\% & 49.32\% & 43.94\%& {\bf 42.84\%}\\
           \hline
           \multirow{2}{*}{CIFAR} & \multirow{2}{*}{2/255} & Clean&42.33\% & 42.15\% & 41.77\% & 42.43\% & 41.86\% & 40.79\%  & {\bf 40.77\%}\\ \cline{3-10}
            & & Verified&55.00\% & 55.11\% & 54.97\% & 54.73\% & 55.07\% & {53.61\%} & {53.61\%}\\
           \hline
    \end{tabular}
\end{table*}
\vspace{-5pt}
\paragraph{Gradient Boosting of Robust Ensemble.} Unlike previous experiments where all base models are given and fixed, we follow Eq.~\eqref{eq:gb_loss} and train models incrementally. We use cross-entropy loss and train an ensemble of 5 models on MNIST with $\epsilon=0.3$ and CIFAR with $\epsilon=\frac{2}{255}$ in Figure~\ref{fig:grad_boost}. We observe that usually the first 2 or 3 models decrease verified errors most (on MNIST from 39.8\% to 34.6\%, and on CIFAR from 51.18\% to 49.55\%. It is challenging to further decrease this error with more models, but gradient boosting allows us to fix the data points lacking of robustness on previous models, achieving better performance than Table~\ref{tb:robust_result}.

\iftoggle{LONG_VERSION}{
\begin{figure}[htbp]
     \centering     
     \begin{subfigure}[b]{0.35\textwidth}
         \centering
         \includegraphics[width=\textwidth]{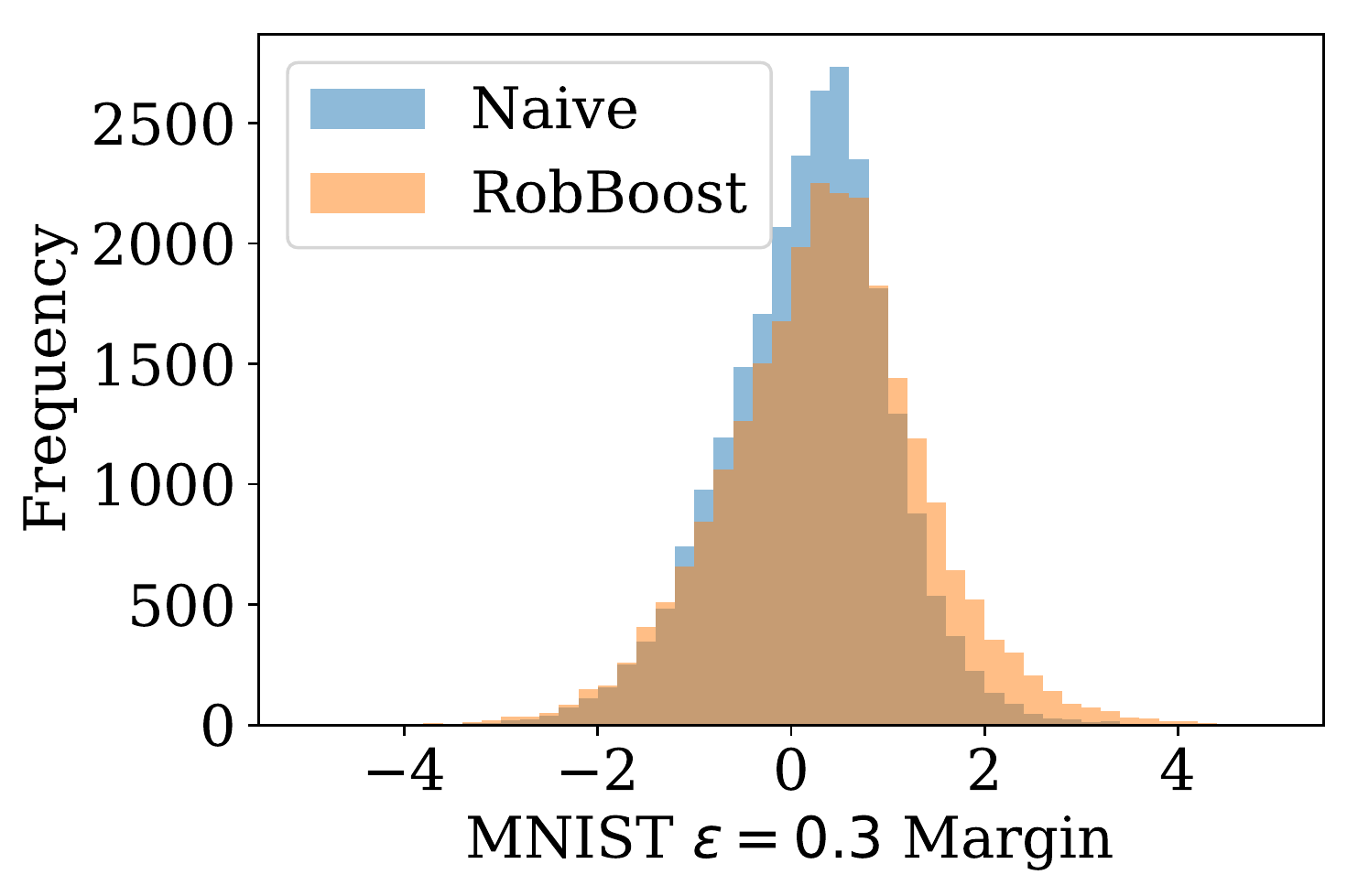}
     \end{subfigure}
    ~
     \begin{subfigure}[b]{0.35\textwidth}
         \centering
         \includegraphics[width=\textwidth]{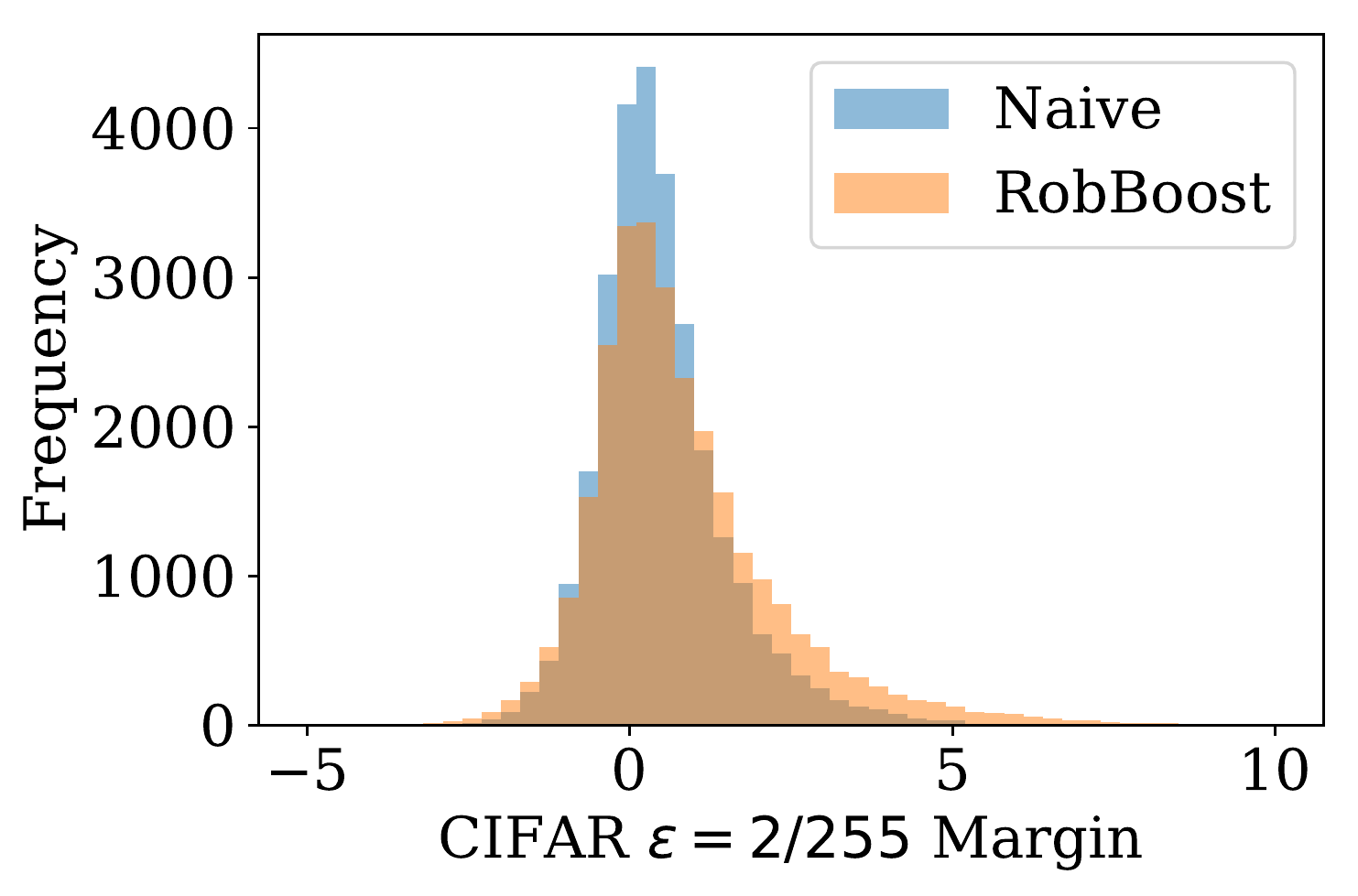}
     \end{subfigure}
     \caption{Distribution of margins' lower bounds. RobBoost moves the margins towards the positive side; a margin greater than 0 indicates that an example is certifiably robust.}  \label{fig:margin_distribution}
\end{figure}
}{}

\iftoggle{LONG_VERSION}{}{\vspace{-15pt}}
\section{Conclusion}
\vspace{-5pt}
We propose the the first ensemble algorithm, RobBoost, to enhance \textit{provable} model robustness by optimally weighting each base model. Our algorithm involves optimizing a surrogate of the lower bound of classification margin through the proposed coordinate descent algorithm, and consistently outperforms a naive averaging ensemble as well as a state-of-the-art single model certified defense in verified error and clean accuracy.

\iftoggle{LONG_VERSION}{
\bibliography{robust_ensemble.bib}
}{
\bibliography{robust_ensemble_short.bib}

\begin{thebibliography}{}

\bibitem[Abbasi and Gagn{\'e}, 2017]{abbasi2017robustness}
Abbasi, M. and Gagn{\'e}, C. (2017).
\newblock Robustness to adversarial examples through an ensemble of
  specialists.
\newblock {\em arXiv preprint arXiv:1702.06856}.

\bibitem[Athalye et~al., 2018]{athalye2018obfuscated}
Athalye, A., Carlini, N., and Wagner, D. (2018).
\newblock Obfuscated gradients give a false sense of security: Circumventing
  defenses to adversarial examples.
\newblock {\em International Conference on Machine Learning (ICML)}.

\bibitem[Carlini and Wagner, 2017a]{carlini2017adversarial}
Carlini, N. and Wagner, D. (2017a).
\newblock Adversarial examples are not easily detected: Bypassing ten detection
  methods.
\newblock In {\em Proceedings of the 10th ACM Workshop on Artificial
  Intelligence and Security}, pages 3--14. ACM.

\bibitem[Carlini and Wagner, 2017b]{carlini2017magnet}
Carlini, N. and Wagner, D. (2017b).
\newblock Magnet and" efficient defenses against adversarial attacks" are not
  robust to adversarial examples.
\newblock {\em arXiv preprint arXiv:1711.08478}.

\bibitem[Carlini and Wagner, 2017c]{carlini2017towards}
Carlini, N. and Wagner, D. (2017c).
\newblock Towards evaluating the robustness of neural networks.
\newblock In {\em 2017 38th IEEE Symposium on Security and Privacy (SP)}, pages
  39--57. IEEE.

\bibitem[Dvijotham et~al., 2018a]{dvijotham18verification}
Dvijotham, K., Garnelo, M., Fawzi, A., and Kohli, P. (2018a).
\newblock Verification of deep probabilistic models.
\newblock {\em CoRR}, abs/1812.02795.

\bibitem[Dvijotham et~al., 2018b]{dvijotham2018training}
Dvijotham, K., Gowal, S., Stanforth, R., Arandjelovic, R., O'Donoghue, B.,
  Uesato, J., and Kohli, P. (2018b).
\newblock Training verified learners with learned verifiers.
\newblock {\em arXiv preprint arXiv:1805.10265}.

\bibitem[Dvijotham et~al., 2018c]{dvijotham2018dual}
Dvijotham, K., Stanforth, R., Gowal, S., Mann, T., and Kohli, P. (2018c).
\newblock A dual approach to scalable verification of deep networks.
\newblock {\em UAI}.

\bibitem[Dvijotham et~al., 2019]{dvijothamefficient2019}
Dvijotham, K.~D., Stanforth, R., Gowal, S., Qin, C., De, S., and Kohli, P.
  (2019).
\newblock Efficient neural network verification with exactness
  characterization.
\newblock {\em UAI}.

\bibitem[Ehlers, 2017]{ehlers2017formal}
Ehlers, R. (2017).
\newblock Formal verification of piece-wise linear feed-forward neural
  networks.
\newblock In {\em International Symposium on Automated Technology for
  Verification and Analysis}, pages 269--286. Springer.

\bibitem[Freund and Schapire, 1997]{freund1997decision}
Freund, Y. and Schapire, R.~E. (1997).
\newblock A decision-theoretic generalization of on-line learning and an
  application to boosting.
\newblock {\em Journal of computer and system sciences}, 55(1):119--139.

\bibitem[Friedman et~al., 2000]{friedman2000additive}
Friedman, J., Hastie, T., Tibshirani, R., et~al. (2000).
\newblock Additive logistic regression: a statistical view of boosting (with
  discussion and a rejoinder by the authors).
\newblock {\em The annals of statistics}, 28(2):337--407.

\bibitem[Friedman, 2001]{friedman2001greedy}
Friedman, J.~H. (2001).
\newblock Greedy function approximation: a gradient boosting machine.
\newblock {\em Annals of statistics}, pages 1189--1232.

\bibitem[Friedman, 2002]{friedman2002stochastic}
Friedman, J.~H. (2002).
\newblock Stochastic gradient boosting.
\newblock {\em Computational Statistics \& Data Analysis}, 38(4):367--378.

\bibitem[Gehr et~al., 2018]{gehr2018ai}
Gehr, T., Mirman, M., Drachsler-Cohen, D., Tsankov, P., Chaudhuri, S., and
  Vechev, M. (2018).
\newblock {AI} 2: Safety and robustness certification of neural networks with
  abstract interpretation.
\newblock In {\em 2018 IEEE Symposium on Security and Privacy (SP)}.

\bibitem[He et~al., 2017]{he2017adversarial}
He, W., Wei, J., Chen, X., Carlini, N., and Song, D. (2017).
\newblock Adversarial example defenses: ensembles of weak defenses are not
  strong.
\newblock In {\em Proceedings of the 11th USENIX Conference on Offensive
  Technologies}, pages 15--15. USENIX Association.

\bibitem[Hein and Andriushchenko, 2017]{hein2017formal}
Hein, M. and Andriushchenko, M. (2017).
\newblock Formal guarantees on the robustness of a classifier against
  adversarial manipulation.
\newblock In {\em Advances in Neural Information Processing Systems (NIPS)},
  pages 2266--2276.

\bibitem[Hosseini et~al., 2019]{hosseini2019dropping}
Hosseini, H., Kannan, S., and Poovendran, R. (2019).
\newblock Dropping pixels for adversarial robustness.
\newblock {\em arXiv preprint arXiv:1905.00180}.

\bibitem[Kariyappa and Qureshi, 2019]{kariyappa2019improving}
Kariyappa, S. and Qureshi, M.~K. (2019).
\newblock Improving adversarial robustness of ensembles with diversity
  training.
\newblock {\em arXiv preprint arXiv:1901.09981}.

\bibitem[Katz et~al., 2017]{katz2017reluplex}
Katz, G., Barrett, C., Dill, D.~L., Julian, K., and Kochenderfer, M.~J. (2017).
\newblock Reluplex: An efficient smt solver for verifying deep neural networks.
\newblock In {\em International Conference on Computer Aided Verification},
  pages 97--117. Springer.

\bibitem[Liu et~al., 2019]{liu2019algorithms}
Liu, C., Arnon, T., Lazarus, C., Barrett, C., and Kochenderfer, M.~J. (2019).
\newblock Algorithms for verifying deep neural networks.
\newblock {\em arXiv preprint arXiv:1903.06758}.

\bibitem[Liu et~al., 2018]{liu2018towards}
Liu, X., Cheng, M., Zhang, H., and Hsieh, C.-J. (2018).
\newblock Towards robust neural networks via random self-ensemble.
\newblock In {\em European Conference on Computer Vision}, pages 381--397.
  Springer.

\bibitem[Madry et~al., 2018]{madry2018towards}
Madry, A., Makelov, A., Schmidt, L., Tsipras, D., and Vladu, A. (2018).
\newblock Towards deep learning models resistant to adversarial attacks.
\newblock In {\em International Conference on Learning Representations}.

\bibitem[Mirman et~al., 2018]{mirman2018differentiable}
Mirman, M., Gehr, T., and Vechev, M. (2018).
\newblock Differentiable abstract interpretation for provably robust neural
  networks.
\newblock In {\em International Conference on Machine Learning}, pages
  3575--3583.

\bibitem[Pang et~al., 2019]{pang2019improving}
Pang, T., Xu, K., Du, C., Chen, N., and Zhu, J. (2019).
\newblock Improving adversarial robustness via promoting ensemble diversity.
\newblock {\em arXiv preprint arXiv:1901.08846}.

\bibitem[Qin et~al., 2019]{qin2018verification}
Qin, C., Dvijotham, K.~D., O'Donoghue, B., Bunel, R., Stanforth, R., Gowal, S.,
  Uesato, J., Swirszcz, G., and Kohli, P. (2019).
\newblock Verification of non-linear specifications for neural networks.
\newblock {\em ICLR}.

\bibitem[Raghunathan et~al., 2018a]{raghunathan2018certified}
Raghunathan, A., Steinhardt, J., and Liang, P. (2018a).
\newblock Certified defenses against adversarial examples.
\newblock {\em International Conference on Learning Representations (ICLR),
  arXiv preprint arXiv:1801.09344}.

\bibitem[Raghunathan et~al., 2018b]{raghunathan2018semidefinite}
Raghunathan, A., Steinhardt, J., and Liang, P.~S. (2018b).
\newblock Semidefinite relaxations for certifying robustness to adversarial
  examples.
\newblock In {\em Advances in Neural Information Processing Systems}, pages
  10900--10910.

\bibitem[Salman et~al., 2019]{salman2019convex}
Salman, H., Yang, G., Zhang, H., Hsieh, C.-J., and Zhang, P. (2019).
\newblock A convex relaxation barrier to tight robust verification of neural
  networks.
\newblock {\em arXiv preprint arXiv:1902.08722}.

\bibitem[Singh et~al., 2018]{singh2018fast}
Singh, G., Gehr, T., Mirman, M., P{\"u}schel, M., and Vechev, M. (2018).
\newblock Fast and effective robustness certification.
\newblock In {\em Advances in Neural Information Processing Systems}, pages
  10825--10836.

\bibitem[Singh et~al., 2019a]{singh2019abstract}
Singh, G., Gehr, T., P{\"u}schel, M., and Vechev, M. (2019a).
\newblock An abstract domain for certifying neural networks.
\newblock {\em Proceedings of the ACM on Programming Languages}, 3(POPL):41.

\bibitem[Singh et~al., 2019b]{Singh2019robustness}
Singh, G., Gehr, T., Püschel, M., and Vechev, M. (2019b).
\newblock Robustness certification with refinement.
\newblock {\em ICLR}.

\bibitem[Sinha et~al., 2018]{sinha2018certifying}
Sinha, A., Namkoong, H., and Duchi, J. (2018).
\newblock Certifying some distributional robustness with principled adversarial
  training.
\newblock In {\em ICLR}.

\bibitem[Strauss et~al., 2017]{strauss2017ensemble}
Strauss, T., Hanselmann, M., Junginger, A., and Ulmer, H. (2017).
\newblock Ensemble methods as a defense to adversarial perturbations against
  deep neural networks.
\newblock {\em arXiv preprint arXiv:1709.03423}.

\bibitem[Tjeng et~al., 2019]{tjeng2017evaluating}
Tjeng, V., Xiao, K., and Tedrake, R. (2019).
\newblock Evaluating robustness of neural networks with mixed integer
  programming.
\newblock {\em ICLR}.

\bibitem[Uesato et~al., 2018]{uesato2018adversarial}
Uesato, J., O'Donoghue, B., Oord, A. v.~d., and Kohli, P. (2018).
\newblock Adversarial risk and the dangers of evaluating against weak attacks.
\newblock {\em International Conference on Machine Learning (ICML)}.

\bibitem[Wang et~al., 2018a]{wang2018mixtrain}
Wang, S., Chen, Y., Abdou, A., and Jana, S. (2018a).
\newblock Mixtrain: Scalable training of formally robust neural networks.
\newblock {\em arXiv preprint arXiv:1811.02625}.

\bibitem[Wang et~al., 2018b]{wang2018efficient}
Wang, S., Pei, K., Whitehouse, J., Yang, J., and Jana, S. (2018b).
\newblock Efficient formal safety analysis of neural networks.
\newblock In {\em Advances in Neural Information Processing Systems}, pages
  6369--6379.

\bibitem[Weng et~al., 2018]{weng2018towards}
Weng, T.-W., Zhang, H., Chen, H., Song, Z., Hsieh, C.-J., Boning, D., Dhillon,
  I.~S., and Daniel, L. (2018).
\newblock Towards fast computation of certified robustness for {ReLU} networks.
\newblock In {\em International Conference on Machine Learning}.

\bibitem[Wong and Kolter, 2018]{wong2018provable}
Wong, E. and Kolter, Z. (2018).
\newblock Provable defenses against adversarial examples via the convex outer
  adversarial polytope.
\newblock In {\em International Conference on Machine Learning (ICML)}, pages
  5283--5292.

\bibitem[Wong et~al., 2018]{wong2018scaling}
Wong, E., Schmidt, F., Metzen, J.~H., and Kolter, J.~Z. (2018).
\newblock Scaling provable adversarial defenses.
\newblock {\em Advances in Neural Information Processing Systems (NIPS)}.

\bibitem[Xiao et~al., 2019]{xiao2018training}
Xiao, K.~Y., Tjeng, V., Shafiullah, N.~M., and Madry, A. (2019).
\newblock Training for faster adversarial robustness verification via inducing
  relu stability.
\newblock {\em ICLR}.

\bibitem[Zhang et~al., 2018]{zhang2018crown}
Zhang, H., Weng, T.-W., Chen, P.-Y., Hsieh, C.-J., and Daniel, L. (2018).
\newblock Efficient neural network robustness certification with general
  activation functions.
\newblock In {\em Advances in Neural Information Processing Systems (NIPS)}.

\bibitem[Zhang et~al., 2019]{zhang2018recurjac}
Zhang, H., Zhang, P., and Hsieh, C.-J. (2019).
\newblock Recurjac: An efficient recursive algorithm for bounding jacobian
  matrix of neural networks and its applications.
\newblock {\em AAAI Conference on Artificial Intelligence}.

\end{thebibliography}
}

\newpage
\appendix
\onecolumn

\section{The Coordinate Descent Algorithm for RobBoost}

\begin{algorithm}[htb]
  \caption{One-step Coordinate Decent Update}
  \label{alg:coord_hinge}
\begin{algorithmic}[1]
\REQUIRE{$N\times (C-1)$ vectors $\bm{\omega}^{j,k} \in \R^{d}$, $\bm{\nu}^{j,k} \in \R^{d}$ and $N\times (C-1)$ scalars $\gamma^{j,k}$ and $\mu^{j,k}$ which are formed based \eqref{eq:coord_rescale}, $j \in [C-1], k \in [N]$, current weight vector $\balp$ and the coordinate $t$ to be updated}
\ENSURE{an updated weight $\alpha_t$ for coordinate $t$}

\FOR{$(j,k) \in [C-1] \times [N]$}
\STATE Solve $r^{j,k}_1, \cdots, r^{j,k}_{d}$, where $\alpha_t \bm{\omega}^{j,k}_l + \bm{\nu}^{j,k}_l = 0, \enskip l \in [d]$
\STATE Sort $r^{j,k}_1, \cdots, r^{j,k}_{d}$ in ascending order $\pi^{j,k}$
\STATE Compute $J^{j,k}_+$, $J^{j,k}_-$ according to~\eqref{eq:coord_j0+} and~\eqref{eq:coord_j0-}
\STATE Compute $a^{j,k}_0$, $b^{j,k}_0$ according to~\eqref{eq:coord_a0} and~\eqref{eq:coord_b0}
\STATE Initial $N \times (C-1)$ index counters $m^{j,k} \leftarrow 0$
\ENDFOR

\STATE Merge sort all $N \times (C-1)$ lists $r^{j,k}$ into one list $r$ with $N \times (C-1)\times d$ elements in ascending order $\pi$
\STATE Maintain two mappings $p(i) \rightarrow (j,k)$, $q(i) \rightarrow l$ where $r_i$ comes from the element $r^{j,k}_l$ during merge
\STATE Initial list $h^{j,k}$ where $h^{j,k} \leftarrow \frac{-b^{j,k}_0}{a^{j,k}_0}$ if $\frac{-b^{j,k}_0}{a^{j,k}_0} > 0$; $\infty$ otherwise \label{line:init_zero}
\COMMENT{stores all the positive 0-crossing points}
\STATE $L_b \leftarrow \sum_{k \in [N]}\sum_{j \in [C-1]} \max (b^{j,k}_0, 0)$, $\alpha_{\text{best}} \leftarrow 0$
\COMMENT{Initial best loss, best $\alpha_t$ at point $\alpha_t=0$}
\STATE $i \leftarrow 1$
\WHILE{$i \leq N \times (C-1)\times d$}

\IF {$r_{\pi(i)} \leq 0$ or $r_{\pi(i)} \geq 1$}
    \STATE $(j, k) \leftarrow p(i)$, $M \leftarrow m^{j,k} + 1$, $m^{j,k} \leftarrow m^{j,k} + 1$
    \STATE {$a^{j,k}_M \leftarrow a^{j,k}_{M-1}$}
    \COMMENT{Skip all $r_{\pi(i)}$ out of the $[0,1]$ range}
    \STATE {$b^{j,k}_M \leftarrow b^{j,k}_{M-1}$}
    \STATE {\bf continue}
\ENDIF

$(j', k') \leftarrow \argmin_{j,k} h^{j,k}$
\IF{$r_{\pi(i)}$ > $h^{j',k'}$ and $h^{j',k'} > 0$}
    \STATE $x \leftarrow h^{j',k'}$
    \COMMENT{Evaluate at a 0-crossing point for term $(j', k')$}
    \STATE $h^{j',k'} \leftarrow \infty$ \COMMENT{Note that we do not increment counter $i$ in this case}
\ELSE
    \STATE $(j, k) \leftarrow p(i)$, $M \leftarrow m^{j,k} + 1$, $m^{j,k} \leftarrow m^{j,k} + 1$
    \STATE{Update $a^{j,k}_M$, $b^{j,k}_M$ recursively using $a^{j,k}_{M-1}, b^{j,k}_{M-1}$} based on~\eqref{eq:update_a} and~\eqref{eq:update_b}
    \IF {$\frac{-b^{j,k}_M}{a^{j,k}_M} > r_{\pi(i)}$}
    \STATE $h^{j,k} \leftarrow \frac{-b^{j,k}_M}{a^{j,k}_M}$
    \COMMENT{Update the 0-crossing point we might encounter}
    \ELSE
    \STATE $h^{j,k} \leftarrow \infty$
    \COMMENT{This piece does not cross 0 when we scan forward}
    \ENDIF
    \STATE $x \leftarrow r_{\pi(i)}$
    \STATE $i \leftarrow i + 1$
\ENDIF

\STATE {Compute $g^{j,k}(x) = a^{j,k}_{m^{j,k}} x + b^{j,k}_{m^{j,k}}$, for all $(j,k) \in [C-1] \times [N]$}

\STATE {Compute Loss at current point $L=\sum_{k \in [N]}\sum_{j \in [C-1]} \max (g^{j,k}(x), 0)$}
\IF{$L < L_b$}  
\STATE {$L_b \leftarrow L$, \enskip $\alpha_{\text{best}} \leftarrow x$}
\COMMENT{Update best loss}
\ENDIF
\ENDWHILE


\STATE {\bf return $\alpha_{\text{best}}$}
\end{algorithmic}
\end{algorithm}

\iftoggle{LONG_VERSION}{
Here we present our full algorithm to optimally update one coordinate in coordinate descent. First, we rewrite our objective function:
\begin{align}
g(\alpha_t) &= \sum_{k \in [N]} \sum_{j \in [C-1]} \max(\|  \alpha_t \bm{\omega}^{j,k} + \bm{\nu}^{j,k}  \|_1 + \gamma^{j,k} \alpha_t + \mu^{j,k} + 1, 0) \\
&\coloneqq \sum_{k \in [N]} \sum_{j \in [C-1]} \max(g^{j,k}(\alpha_t) ,0) \label{eq:coord_rescale_obj}
\end{align}
}{}

There are $N \times (C-1)$ terms in summation~\iftoggle{LONG_VERSION}{\eqref{eq:coord_rescale_obj}}{\eqref{eq:coord_rescale}}, and we need to maintain ``effective slope'' $a^{j,k}$ and ``effective intercept'' $b^{j,k}$ for each term $(j,k)$ using the technique presented in the main text. Thus most variables have superscript $(j,k)$. Note that to deal with the $\max$ function in hinge loss, for each term we need to dynamically maintain the possible zero crossing point, which is stored in $h^{j,k}$ (line 9). If we reached a zero-crossing point before any other $r_{\pi{(i)}}$, we need to evaluate the function value at this zero-crossing point first (line 19). Additionally, once the effective slope $a^{j,k}_M$ and effective intercept $b^{j,k}_M$ change due to a sign change in absolute value terms, we need to update the zero-crossing point $h^{j,k}$ (line 25).
\iftoggle{LONG_VERSION}{
In each iteration of coordinate descent, we choose a coordinate $t$,  obtain~\eqref{eq:coord_opt} using Algorithm~\ref{alg:coord_hinge} and set $\alpha_t \leftarrow \alpha_t^*$. Initially, we can set $\alpha_t = \frac{1}{T}$, then we can run several epochs of coordinate descent; in each epoch we optimize over all $\alpha_t$ once, either in a cyclic order or a random order. Although coordinate descent cannot always find the global optimal solution, we found that it usually reduces the loss function sufficiently and rapidly; 2 or 3 epochs are sufficient for all our experiments.
}{}





\iftoggle{LONG_VERSION}{
\section{Model Details}
\label{sec:models}
We implement all our models using PyTorch.
Using model structures detailed in Table~\ref{tb:models}, we obtain 12 MNIST models and 11 CIFAR-10 models (without the last largest ResNet due to out of memory), robustly trained using the method proposed in \cite{wong2018scaling}. For MNIST and CIFAR-10 we use the same model structure; only the input image shape differs.

For experiments on ensemble of robustly trained models, we use all available models. For experiments on feature subsample, we use model K for both MNIST and CIFAR-10 datasets. For experiments on the ensemble of graident boosted models, we use model A,C,E,G,K for MNIST and B,G,I,J,K for CIFAR-10 dataset.

\begin{table*}[bh]
\caption{Models structures. All layers are followed by ReLU activations. The last fully connected layer with a fixed 10-dimensional output is omitted. ``Conv $k w\times h+s$'' corresponds to a 2D convolutional layer with $k$ filters of size $w\times h$ using a stride of $s$ in both dimensions. $[\cdot]$ stands for a Resnet block.}
\label{tb:models}
\vskip 0.15in
\begin{center}
\begin{adjustbox}{max width=\textwidth}
\begin{tabular}{l|l}
\toprule
Name & Model details  \\
\midrule
A & Conv 4 $4 \times 4$+2, Conv 8 $4 \times 4$+2, FC 128 \\ 
B & Conv 8 $4 \times 4$+2, Conv 16 $4 \times 4$+2, FC 256 \\
C & Conv 4 $3 \times 3$+1, Conv 8 $3 \times 3$+1, Conv 8 $4 \times 4$+4, FC 64 \\
D & Conv 8 $3 \times 3$+1, Conv 16 $3 \times 3$+1, Conv 16 $4 \times 4$+4, FC 128 \\
E & Conv 4 $5 \times 5$+1, Conv 8 $5 \times 5$+1, Conv 8 $5 \times 5$+4, FC 64 \\
F & Conv 8 $5 \times 5$+1, Conv 16 $5 \times 5$+1, Conv 16 $5 \times 5$+4, FC 128 \\
G & Conv 4 $3 \times 3$+1, Conv 4 $4 \times 4$+2, Conv 8 $3 \times 3$+1, Conv 8 $4 \times 4$+2, FC 256, FC 256 \\
H & Conv 4 $3 \times 3$+1, Conv 4 $4 \times 4$+2, Conv 8 $3 \times 3$+1, Conv 8 $4 \times 4$+2, FC 512, FC 512 \\
I & Conv 8 $3 \times 3$+1, Conv 8 $4 \times 4$+2, Conv 16 $3 \times 3$+1, Conv 16 $4 \times 4$+2, FC 256, FC 256 \\
J & Conv 8 $3 \times 3$+1, Conv 8 $4 \times 4$+2, Conv 16 $3 \times 3$+1, Conv 16 $4 \times 4$+2, FC 512, FC 512 \\
K & Conv 4 $3 \times 3$+1, [Conv 8 $4 \times 4$+2, Conv 8 $3 \times 3$+1 Conv 8 $3 \times 3$+1], [Conv 8 $4 \times 4$+2, Conv 8 $3 \times 3$+1 Conv 8 $3 \times 3$+1], FC 128\\
L & Conv 8 $3 \times 3$+1, [Conv 16 $4 \times 4$+2, Conv16 $3 \times 3$+1 Conv 16 $3 \times 3$+1], [Conv 16 $4 \times 4$+2, Conv 16 $3 \times 3$+1 Conv 16 $3 \times 3$+1], FC 256\\
\bottomrule
\end{tabular}
\end{adjustbox}
\end{center}
\vskip -0.1in
\end{table*}
}{}

\end{document}